\pdfoutput=1

\documentclass[11pt]{article}

\usepackage[]{acl}

\usepackage{times}
\usepackage{latexsym}

\usepackage[T1]{fontenc}

\usepackage[utf8]{inputenc}

\usepackage{microtype}

\usepackage{graphicx}
\usepackage{amsmath}
\usepackage{booktabs}
\usepackage{multirow}
\usepackage{hyperref}
\usepackage{comment}
\usepackage{xspace}
\usepackage[colorinlistoftodos]{todonotes}
\usepackage{enumitem}
\usepackage{microtype}
\usepackage{float}
\usepackage{natbib}

\newcommand{\CSadj}{commonsense\xspace}
\newcommand{\CSn}{common sense\xspace}

\newcommand{\CN}{ConceptNet\xspace}

\newcommand{\QT}[1]{``{#1}''\xspace}
\newcommand{\MCQ}{{precondition}\xspace}
\newcommand{\MCQs}{{preconditions}\xspace}

\newcommand{\CQ}{\textit{PaCo}\xspace}
\newcommand{\paco}{\textit{PaCo}\xspace}

\newcommand{\mcqa}{P-MCQA\xspace}
\newcommand{\nli}{P-NLI\xspace}
\newcommand{\gen}{P-G\xspace}

\newcommand{\fact}{statement\xspace}
\newcommand{\facts}{statements\xspace}
\newcommand{\Fact}{Statement\xspace}

\definecolor{USCgold}{HTML}{F6C400}
\newcommand{\highlight}[1]{{\setlength{\fboxsep}{2pt}\hspace{-2pt}\colorbox{USCgold}{#1}}}

\title{PaCo: Preconditions Attributed to Commonsense Knowledge}

\author{
Ehsan Qasemi{$^{*\dagger}$} 
    \and Filip Ilievski{$^\dagger$} 
    \and Muhao Chen{$^{*\dagger}$} 
    \and Pedro Szekely{$^{*\dagger}$}
    \\
    \textbf{$*$}Department of Computer Science, University of Southern California
    \\
    \textbf{$\dagger$}Information Sciences Institute, University of Southern California
    \\ 
    \{qasemi,muhaoche,szekely\}@usc.edu
    , 
    \{ilievski\}@isi.edu
\\}

\begin{document}
\maketitle

\begin{abstract}
	Humans can seamlessly reason with circumstantial preconditions of commonsense knowledge. We understand that \textit{a glass is used for drinking water}, unless the glass is broken or the water is toxic.
	Despite state-of-the-art (SOTA) language models' (LMs) impressive performance on inferring commonsense knowledge, it is unclear whether they understand the circumstantial \MCQs. 
	To address this gap, we propose a novel challenge of reasoning with circumstantial \MCQs. We collect a dataset, called \CQ, consisting of $12.4$ thousand preconditions  of \CSadj \facts expressed in natural language.
	Based on this dataset, we create three canonical evaluation tasks and use them to examine the capability of existing LMs to understand situational \MCQs.
	Our results reveal a $10$-$30\%$ gap between machine and human performance on our tasks, which shows that reasoning with preconditions is an open challenge.\footnote{Code and data on \hyperlink{https://github.com/luka-group/PaCo}{https://github.com/luka-group/PaCo}}
\end{abstract}
	\section{Introduction}
\label{sec:intro}
Improving a system's ability to reason with \CSadj knowledge is at the frontier of natural language processing (NLP) research, as a critical component in many knowledge-driven tasks such as question answering~\cite{wang2019superglue,talmor2018commonsenseqa}, machine reading comprehension~\cite{sakaguchi2020winogrande}, narrative cloze~\cite{mostafazadeh-etal-2016-corpus}, and dialogue systems~\cite{adiwardana2020towards,young2018augmenting}.
Recently, dozens of systems~\cite{raffel2019T5,khashabi2020unifiedqa,liu2019roberta,devlin2018bert} and learning resources~\cite{sap2019socialiqa,mostafazadeh2020glucose,rudinger2020thinking,Bhagavatula2020Abductive} have been proposed, focusing on various aspects of \CSadj knowledge such as naive physics and naive psychology.

In cognitive studies, the \emph{theory of affordance}~\cite{affordance,chemero2003outline} suggests that understanding the circumstances in which an action or \fact is possible or impossible is a key aspect of human intelligence.
For example, a glass may be used for drinking water, under an implicit assumption that the water is at normal temperature, but may not if the glass is shattered.
Accordingly, we argue that for an NLP reasoner to understand \CSn, it should comprehend the contextual \textit{\MCQs} associated with commonsense \facts.
Such contextual \MCQs can naturally be categorized into two classes: the ones that \emph{enable} the \facts, and the ones that \emph{disable} them \cite{fikes1971strips, hobbs2005toward}.

\begin{figure}[t]
    \centering
    \includegraphics[width=\linewidth]{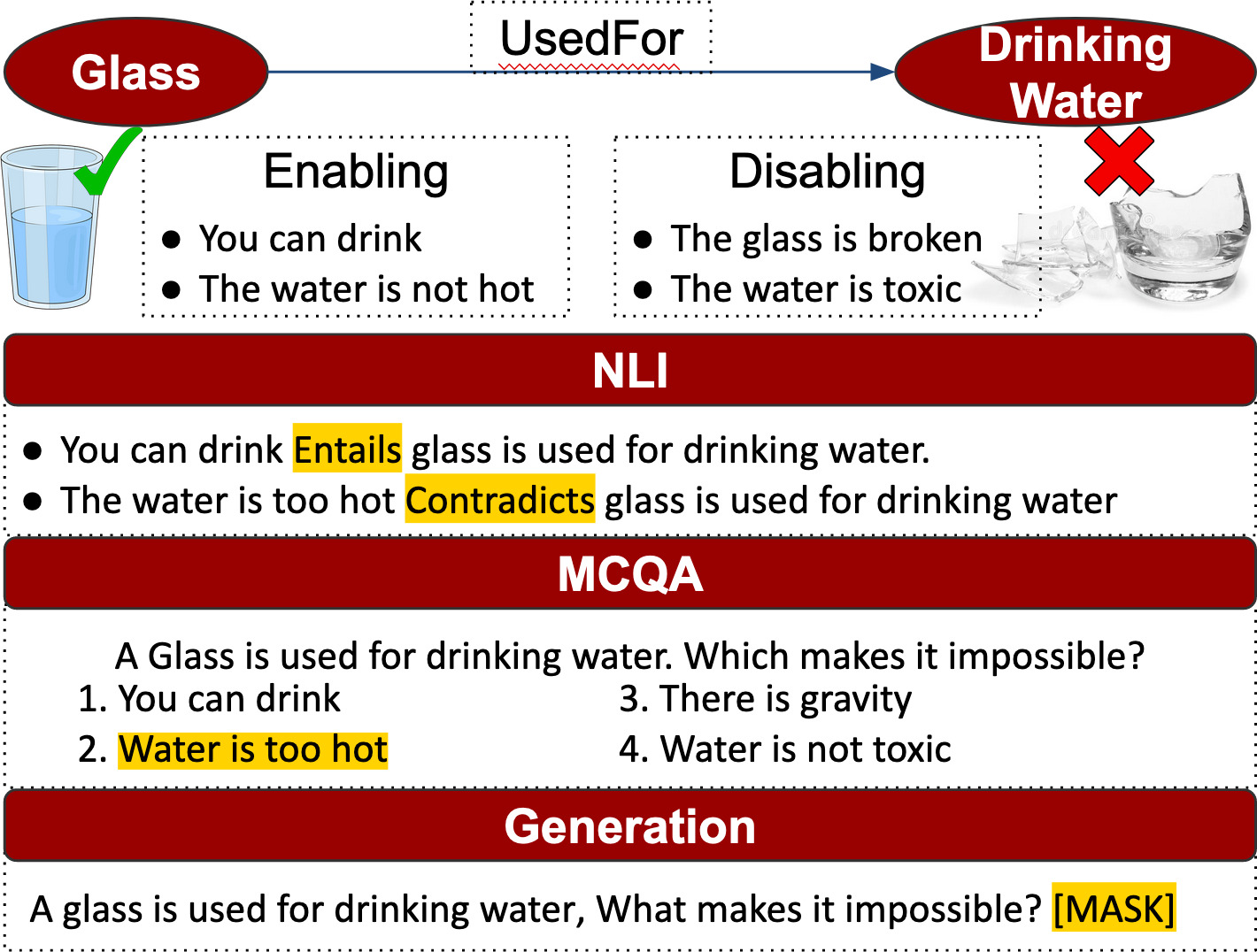}
    \caption{Overview of the \CQ data collection and instances of the three tasks derived from it.}\label{fig:mcq_overview}
    \vspace{-1em}
\end{figure}

Causal preconditions may be partially inferred from text~\cite{mostafazadeh2020glucose,kwon-etal-2020-modeling}, however: 1) as is the case in many other aspects of \CSn, we rarely write them explicitly in our text; 2) when mentioned in the text, it is difficult for models to distinguish whether they represent causation or correlation.
Similar to our work, \citet{rudinger2020thinking} collect the \MCQs by crowdsourcing. 
Here, the \MCQs are seen as soft assumptions, namely: \emph{weakeners} and \emph{strengtheners}, which provides a model only with the relative correlation between \facts, and is not explicitly testing the model on the underlying \MCQs of the \fact.
Instead, we propose to define the problem based on the crisp conditioning of \emph{disablers} and \emph{enablers}, which forces the LM to learn the decisive \MCQs of a \fact and facilitates explainability based on them.
In comparison to a hard logical connection modeled 
by the crisp condition, although the notion of \emph{weakener} is also helpful to the commonsense reasoner, it raises additional questions like “by how much?”, or “is the statement still valid?”. 
Whereas in the notion of \emph{disablers}, even though annotations are more difficult to collect, it can at least take the system one step forward by sorting out the clutter of the irrelevant \facts.


This paper presents a systematic study on the problem of situational \MCQs expressed in natural language.
As the \textbf{first} contribution, we define a new problem of reasoning with \emph{enabling} and \emph{disabling} \MCQs associated with \CSadj \facts (Section~\ref{sec:preconditions}).
Given a \fact, the task is to infer the \MCQs that make the \fact possible~(\emph{enabling}) or impossible~(\emph{disabling}). 
Understanding such preconditions of commonsense knowledge would enable reasoning systems relying on a \CSadj knowledge base to decide when to use a given commonsense \fact.
For example, given the \fact \QT{Glass is used for drinking water} in \CN~\cite{speer2016conceptnet}, a system should know that it is only possible if the \QT{water is not too hot}, and it is impossible when \QT{the water is toxic}.

To foster research on preconditions of commonsense knowledge, we develop \CQ, a rich crowdsourced dataset with \emph{enabling} and \emph{disabling} \MCQs of \CSadj \facts (Section~\ref{sec:corequisite}), as the \textbf{second} contribution of this paper. 
For \CQ, we start by extracting available commonsense \facts. 
We then design and execute a crowdsourcing task to gather \MCQs of the \facts by asking participants: \emph{what makes the \fact possible/impossible?} for each of the \facts.
\CQ contains $12.4$K labeled \MCQs ($6.6$K \emph{enabling}, $5.8$K \emph{disabling}), corresponding to $3*1$K edges from three representative relations in \CN~\cite{speer2016conceptnet}, covering knowledge on utility, causality, and motivation.
Example preconditions are illustrated in Fig.~\ref{fig:mcq_overview}.
These tasks for the first time allow analysis beyond what is done in prior work that cover enabling preconditions only. Particularly, they realize a head-to-head comparison of enabling and disabling statements which was not possible before. 
Besides, they allow analysis of the impact of the knowledge types (e.g., utility) on the task difficulty for both humans and neural language models.

Our \textbf{third} contribution is an extensive NLP benchmarking based on \CQ.
To this end, we transform \CQ into three tasks on \textbf{P}reconditions: \textbf{N}atural \textbf{L}anguage \textbf{I}nference (\nli), \textbf{M}ultiple-\textbf{C}hoice \textbf{Q}uestion \textbf{A}nswering (\mcqa), and \textbf{G}eneration (\gen).
The three canonical tasks seek to provide a comprehensive evaluation of the ability of natural language reasoners to understand circumstantial \MCQs (Section~\ref{sec:benchmark_design}).
These three tasks examine the understanding of preconditions of a number of SOTA language models and reasoners, such as DeBERTa~\cite{he2020deberta}, and UnifiedQA~\cite{khashabi2020unifiedqa}.
Results show that SOTA methods largely fall behind human performance,  therefore indicating the need for further research in order to improve the comprehension of contextual \MCQs by \CSadj reasoners (Section~\ref{sec:results}).

\section{Preconditions in Commonsense Reasoning}
\label{sec:preconditions}

\paragraph{Problem Definition.}
Commonsense \facts describe well-known information about concepts, and, as such, they are acceptable by people without need for debate~\cite{sap2019atomic,ilievski2020commonsense}.
A commonsense statement can be formalized as $s=(h,r,t)$, where $h$ and $t$ are head and tail concepts, and $r$ is the relation type. 

Following the notion of \QT{causal complex} \cite{hobbs2005toward}, we define the \MCQ~$P_f$ as a collection of eventualities (events or states) that results in $s$ to happen.
Such \MCQs contain eventualities that either \textit{allow}~($p_{f}^+ \in P_f$) or \textit{prevent}~($p_{f}^- \in P_f$) the \fact to happen. Here, to \textit{prevent} means to \textit{allow} the negation of the \fact~
\cite{fikes1971strips}.
While enumerating a priori all such causal eventualities is impossible, people are still able to reason about them in a given situation~\cite{hobbs2005toward}.
Notably, preconditions are \textit{implicit}, i.e., we usually omit them from conversation as they are considered obvious~\cite{grice1975logic}.
\citet{shoham1990nonmonotonic} and \citet{hobbs2005toward} distinguish between two type of \MCQs, based on causal connections (\emph{hard}), or material implication (tends to cause; \emph{soft}). 
Here we focus on the more restrictive, \emph{hard} \MCQs; for soft \MCQs, see~\cite{rudinger2020thinking}.



In this work, the problem of reasoning with \MCQs is attempted in two ways: discriminative and generative~(cf. Table~\ref{tab:simple-test}).
In the discriminative setting, given a \fact~$f$ and a \MCQ~($p$), a model is expected to infer if the fact is still valid~($p \in P_{f}^+$) or not~($p \in P_{f}^-$).
In the generative setting, given only the \fact~($f$), a model is requested to compose a reasonable disabling~($p_{f}^-$) or enabling~($p_{f}^+$) \MCQ.

\begin{table*}
    \centering
    {\small
    \begin{tabular}{lp{0.50\textwidth}p{0.30\textwidth}}
        \hline \textbf{Model} & \textbf{Input} & \textbf{Output}\\
        \hline
            UnifiedQA
            & A net is used for catching fish. What makes this impossible? (A) You are in water (B) You are in downtown LA
            & You are in water
        \\
            UnifiedQA
            & A net is used for catching fish. What makes this impossible?
            & A net is used for catching fish.
        \\
            GPT2
            & A glass is used for drinking water only if, the glass
            & is covered in a protective coat or can be removed with cold water.
        \\
            GPT2
            & A glass is used for drinking water only if, the water
            & is acidic, not fresh.
        \\
        \hline
    \end{tabular}
    }
    \caption{Test of language model's understanding of preconditions}
    \label{tab:simple-test}
    \vspace{-1em}
\end{table*}

\paragraph{Motivating Examples.}
In a preliminary investigation, we assess the ability of SOTA language models: GPT2~\cite{gpt2}, and UnifiedQA~\cite{khashabi2020unifiedqa}, to reason with preconditions. 
As shown in Table~\ref{tab:simple-test}, both models appear to fall short of reasoning with enabling and disabling factors of commonsense statements, regardless of whether the prompt task form is presented as multiple-choice question answering~(row 1), or as text completion~(rows 2-4). 
This observation is not surprising, considering that reasoning with preconditions is an under-addressed research challenge. 
Yet, it motivates the urgency for this problem to be studied in depth, which is the goal of this paper.

\section{\CQ}
\label{sec:corequisite}
This section introduces the procedure of developing the \CQ dataset.
We start by selecting relevant \CSadj facts~(Section~\ref{subsec:edges}), and crowdsourcing \MCQs for each \fact~(Section~\ref{subsec:datacollection}).
Finally, we present the \CQ data statistics~(Section~\ref{subsec:data-statistics}).

\subsection{Edge Selection}
\label{subsec:edges}

We extracted relevant \CSadj facts from \CN~\cite{speer2016conceptnet}. 
We chose \CN due to its breadth of knowledge and popularity in prior research~\cite{feng2020scalable,lin2019kagnet,ma2019towards-hykas}.
\CN is a publicly available common sense knowledge resource.
It contains 3.4 million English assertions between concepts (e.g.,~\QT{Glass}, \QT{Drinking\_water}, \QT{Person}), and covers a wide range of knowledge types, including spatial, physical, and temporal knowledge, as well as social and cognitive knowledge about everyday situations.

We performed a pilot analysis of different knowledge types in \CN to help us decide which of them were suitable to be annotated with preconditions.
Namely, we sampled $20$ random edges for each relation and checked how well one could annotate them with preconditions.
Our analysis revealed that not all relations lent themselves naturally for annotation with enabling or disabling preconditions.
Specifically, we observed that some relations (e.g., \emph{Related To}) are underspecified in their meanings, and others, like \emph{IsA}, are often truisms.
Our investigation has revealed that it is difficult to come up with \MCQs for these relations.
Furthermore, we observed that some relations, like \emph{CreatedBy}, could be easily annotated with enabling conditions, but not with disabling ones.
The opposite was observed for \emph{PartOf}.

We opted for the relations \emph{UsedFor}, \emph{Causes}, and \emph{Desires}, because of their suitability for annotation of \MCQs, their relatively high number of statements, and their representativeness of three different dimensions of knowledge: utility, temporal, and motivational knowledge~\cite{ilievski2021dimensions}. 
Following the intuition that not all statements can be annotated with \MCQs, e.g., \emph{(Looking through telescope, Usedfor, viewing heavens)},
we computed the correlation between a hand-annotated suitability judgment for the \MCQ statements, and the several quantitative scores: DICE metrics~(\citealt{chalier2020joint}; e.g.,~salience), LM perplexity, and edge weights in \CN.
However, none of these scores had a strong correlation with the suitability for annotating \MCQs (Appendix~\ref{sec:appendix:edge} contains the calculated correlations for \emph{UsedFor}).
Therefore, we opted for the relations \emph{UsedFor}, \emph{Causes}, and \emph{Desires}, because of their suitability for annotation of \MCQs, high number. Also they are representative of three different dimensions of knowledge: utility, temporal, and motivational knowledge~\cite{ilievski2021dimensions}.
We sampled 1K edges from each and lexicalized them into human readable sentences using relation-specific templates (see Appendix~\ref{sec:appendix:lexic}).

\subsection{Data Collection}
\label{subsec:datacollection}

\paragraph{Mechanical Turk}
We used Amazon Mechanical Turk~\cite{crowston2012amazon} to collect data on \MCQs for the lexicalized \facts as part of Institutional Review Boards~(IRB) approved (as exempt) study.
For this, we asked the participants to provide short responses to the question: \QT{What makes the \fact possible/impossible?} for each of the lexicalized statements from \CN.
Due to financial limitations, we restricted our annotations to
3 enabling and 3 disabling judgments for each \fact.
While the goal of \CQ is not to exhaust all possible \MCQs associated with each \fact, for some statements we observed duplicate answers, signaling a near-saturation point.

Further details on the data collection design, including annotator qualification, and survey design details are given in Appendix~\ref{sec:appendix:mturk}.
With this procedure, we collected a total of 18K enabling and disabling \MCQs.

\paragraph{Quality Control}
\label{subsec:quality-control}
We use a mixture of automated and expert annotations for quality control.
The automated quality control consisted of three rules that we can programmatically check: 1) not using negative words like \QT{not}, 2) not using pronouns, and 3) proper sentence lengths. 
In order to measure the informativeness and relevance of the remaining annotations, we use expert annotation. 
Specifically, for a subset of the recorded responses we asked the annotator to classify the response into three categories, each representing a specific level of informativeness in the response:
1) \textit{Truism}: the response is correct, but it is not specific to the situation (e.g., \emph{being broken/functional} or \emph{being available/unavailable});
2) \textit{Informative}: the response is correct and is adding information that is not mentioned in the prompt, while not being a truism~(i.e., is specific); 3) \textit{Irrelevant}: any response that is not placed into the previous two categories.
For \CQ, we remove the answers from the \textit{Irrelevant} category, while truism answers could be removed subsequently if so desired.

\subsection{Dataset Statistics}
\label{subsec:data-statistics}
This data collection procedure resulted in a total of 9k enabling and 9k disabling preconditions for each of the 1k ConceptNet edges selected for \emph{UsedFor}, \emph{Causes}, and \emph{Desires} relations respectively.
After filtering out responses in low quality and those marked as \emph{Invalid} by crowd annotators, \CQ contains $12.4$K annotations ($6.6$K \emph{enabling}, $5.8$K \emph{disabling}).
Our expert annotation on $10\%$ of the 6K annotations with \emph{UsedFor} relation showed that in $93\%$ of the crowdsourced responses are informative, whereas only $5\%$ of the responses are irrelevant.
The quality of the responses is lower for the two other relations: $70\%$ informative responses for \emph{Causes} and $61\%$ for \emph{Desires}.
This shows that the two relations are semantically more challenging to human annotators compared to a utility relation like \emph{UsedFor}.
We also observed that on average it took the annotators $3.5$ times longer to submit a responses for these two relations, which confirms that \textit{UsedFor} is the most suitable of the three relations for associating \MCQs. 


\section{Tasks}
\label{sec:benchmark_design}

Given the data collected in Section~\ref{sec:corequisite}, we devise three complementary tasks to showcase the possible ways one could use the \CQ data to evaluate the current SOTA models' understanding of circumstantial \MCQs.
We select
\textbf{P}reconditions \textbf{N}atural \textbf{L}anguage \textbf{I}nference (\nli) and \textbf{P}reconditions \textbf{M}ultiple-\textbf{C}hoice \textbf{Q}uestion \textbf{A}nswering (\mcqa) as representative \emph{discriminative} tasks, and \textbf{P}reconditions \textbf{G}eneration (\gen) task as a \emph{generative} task.
Table~\ref{tab:benchmark-instances} summarizes the tasks and provides an example for each of them.
In the rest of this section, we describe each task in detail and discuss the steps to prepare it from the raw \MCQ data. 
This preparation is fully automatic, and no human annotation or supervision signals have been used.

\begin{table}
    \small
    \centering
    \begin{tabular}{p{0.16\linewidth}p{0.75\linewidth}}
        \hline \textbf{ID} & \textbf{Instance}
        \\ \hline
        \nli
        & \textit{\underline{Hypothesis}}: \emph{A net is used for catching fish}
        \\
        & \textit{\underline{Premise}}: \emph{We are in a desert}
        \\
        & \textit{\underline{Label}}: \emph{Contradiction}
        \\
        \hline
        \mcqa
        & \textit{\underline{Question}}: \emph{A net is used for catching fish. When is this impossible?}
        \\
        & \textit{\underline{Choices}}: \emph{(A) You are in sea, (B) The boat is moving, (C) Net has a large hole in
        it.}                       \\
        \hline
        \gen
        & \textit{\underline{Question}}: \emph{A net is used for catching fish. When is this impossible?}
        \\
        & \textit{\underline{References}}: \emph{(-) Net has a large hole in it, (-) You are in downtown LA, (-) There
        are no fish in the water} \\
        \hline
    \end{tabular}
    \caption{Example of the three tasks in \CQ.}
    \label{tab:benchmark-instances}
    \vspace{-2em}
\end{table}

\paragraph{\nli Task}
\label{subsec:nli_benchmark}

Natural Language Inference (NLI) refers to tasks where given a sentence pair composed of a \emph{hypothesis} and a \emph{premise}, the system has to decide whether the hypothesis is true (entailment), false (contradiction), or undetermined (neutral) given the premise~\cite{mnli}.
Each of the preconditions (e.g.,~\QT{water is clean} or \QT{water is polluted}) of a \fact can directly serve as a \textit{premise} in the sense of NLI. Enabling preconditions correspond to \textit{entailment} cases (e.g.,~\QT{water is clean} \emph{entails} \QT{water is used for drinking}), whereas disabling preconditions can be annotated as \textit{contradictions} (e.g.~\QT{water is polluted} \emph{contradicts} \QT{water is used for drinking}).
The \nli task consists of $12.4$K entries, with $6.6$K entailment and $5.8$K contradiction cases.

\paragraph{\mcqa Task}
\label{subsec:mcqa_benchmark}

\CQ can also be directly converted to a multiple-choice question answering (MCQA) task in three steps.
First, for each \fact, each enabling (disabling) response is paired with three disabling (enabling) responses from the same statement.
These three responses naturally act as negative samples (distractors), allowing us to have high-quality and fair questions. 
The question of the MCQA instance is then formed by appending \QT{What makes this possible?} or \QT{... impossible?} to the lexicalized \fact.
Second, in order to have more distractors and increase the number of multiple-choice instances
we applied the two negative sampling methods used by~\citet{lyu2020reasoning}: Cosine Similarity Filtering, and Question/Answer Shuffling.
Finally, in order to remove the annotation artifacts from the data, hence trivial instances, and prevent the models to exploit these artifacts instead of answering the questions, we used the \emph{Lite} variation of the Adversarial Filtering method, which has been introduced in~\citet{sakaguchi2020winogrande} and formalized in~\citet{le2020adversarial}.
This resulted in a \mcqa task with $47$K multiple choice questions, each with $4$ choices.

\paragraph{\gen Task}
\label{subsec:PG}
Despite our adversarial strategies, it remains possible that reasoning systems may identify annotation artifacts  \cite{gururangan-etal-2018-annotation} in the data and solve the discriminative tasks without correctly performing the logical inference, as a result of those artifacts~\cite{le2020adversarial}.
Hence, we provide a third formulation as a generative commonsense reasoning task.
In this task, we present the system with the exact question that has been presented to the human annotators, thereby mimicking the human annotation task of writing down the \MCQ as a natural language sentence.
We then evaluate the model's response using the human responses as references.
After removing the low-quality and \emph{Invalid} responses from \CQ, the \gen task consists of $5.2$K instances, with an average of $2.4$ reference sentences per instance.

\section{Experiments}
\label{sec:results}
This section pitches SOTA language models against the three tasks derived from \CQ~(Section~\ref{subsec:base-results}), dives deep into the tuning process to pinpoint time of comprehension~(Section~\ref{sec:curve-results}), investigates how LMs react to different relation types~(Section~\ref{sec:other-ralation-results}), and finally revisits the distinction between soft and hard \MCQs~(Section~\ref{sec:soft-hard-results}).

\subsection{Evaluating SOTA on \CQ Tasks}
\label{subsec:base-results}
We assess our benchmark through evaluating representative NLP systems on the three tasks.
This part starts with details about experimental setups (Section~\ref{subsubsec:base-results-setup}), followed by result analysis for the three tasks (Sections~\ref{subsubsec:task-base-results}).

\subsubsection{Experimental Setup}
\label{subsubsec:base-results-setup}
For each task, we start from available pretrained models and evaluate their performance on the test set in zero-shot and fine-tuned setups.
To create the test set, we use a uniform random split of the \facts that each task's instance is stemed from. 
For the split we use the $[0.45, 0.15, 0.40]$ ratio of the data for train/dev/test.
The rationale for splitting based on the \facts instead of the task instances is to prevent data leakage into the test sets through shared edges.
The experiments are conducted on a commodity workstation with an Intel Xeon Gold 5217 CPU and an NVIDIA RTX 8000 GPU .
For all the tasks, we use \emph{allennlp}~\cite{Gardner2017AllenNLP} library for the Textual Entailment~(TE) model~\cite{parikh2016decomposable} and use \emph{huggingface}~\cite{huggingface} for the rest of them.

For the human evaluations of \nli and \mcqa, we used a small ($100$) sample from test subset of each task and asked a CS graduate student to answer them.
We then report the respective evaluation metric based on the task, as detailed below.

\subsubsection{Evaluation Protocols}
\label{subsubsec:evaluation-protocole}
For \nli, we use \emph{F1-Macro} score on the ground-truth labels and report the results on the unseen test split of the data.

For \mcqa, we evaluate the systems' performance based on their default evaluation protocols as discussed below.
For RoBERTa~\cite{liu2019roberta}, we use the LM coupled with a linear regression layer as classification head.
In this method, the LM is tasked with embedding each question/answer pair, and the classification head assigns a score to the pair.
Later for each MC instance, the question/answer pair with the highest score is selected as the output choice.
We report the accuracy score (code from~\cite{sklearn}) based on the output choices from the model.
For UnifiedQA, we follow the original setting by ~\citet{khashabi2020unifiedqa} to let the model conduct sequence-to-sequence generation based on the question. 
Here, the question and all choices are feed to the model, and it is expected to generate the correct choice's text.
We then report the f1 score by selecting the one that is closest to the generated answer from the candidate choices.

For \gen, to automatically evaluate the machine-generated answers of the models, we use \emph{Bleu-2}~\cite{papineni2002bleu} (code from~\cite{nltk}) and \emph{ROUGE-2}~\cite{lin-2004-rouge} (code from~\cite{huggingface}) metrics. 
We do not use methods with large n-gram match (e.g., \emph{Bleu-4}) for two reasons.
\emph{First}, the small number of reference sentences (at most 3) made most of model's output not matching any reference sentence.
\emph{Second}, relatively short reference sentences leads to no 4-gram match and mostly zero \emph{Bleu-4} scores.

For the human evaluation score of the machine generated responses, we sample $100$ responses and use a method similar to \emph{quality control} method in Section~\ref{subsec:quality-control} (here we consider the \emph{Truism} responses as \emph{Informative}), and report the percentage of \emph{informative} responses from tuned models.

\begin{table}
    \small
    \centering
    \begin{tabular}{lll}
        \hline \textbf{Model} &\textbf{0-Shot} &\textbf{Tuned}\\ \hline
        AllenNLP TE     & 0.34 & 0.85 \\
        RoBERTa-large-MNLI & 0.47 & 0.90 \\
        BART-large-MNLI    & \highlight{0.48} & 0.90 \\
        DeBERTa-base-MNLI & 0.37 & 0.91 \\
        DeBERTa-large-MNLI & 0.36 &\highlight{0.94} \\
        DeBERTa-xl-MNLI & 0.37 & 0.91 \\
        \hline
        Expert Human & 0.99 & - \\
        Random Baseline & 0.5 & - \\
        \hline
    \end{tabular}
    \caption{F1-Macro results of SOTA systems on \nli task based on \CQ. Best values are \highlight{highlighted}.}
    \label{tab:NLI-results}
    \vspace{-1.3em}
\end{table}

\subsubsection{Results and Discussions}
\label{subsubsec:task-base-results}
We hereby separately discuss the performance of SOTA models on the three tasks in details.

\emph{(1) \nli Results}
As shown in Table~\ref{tab:NLI-results}, all systems tend to get near-random results in the zero-shot setup.
In case of the \emph{BART-large-MNLI} model, although the zero-shot \emph{F1-Macro} score is higher, it is far from human-level score ($1.00$).
We observe that even models that are trained on large and diverse learning resources (e.g.~MNLI~\cite{mnli}) are not able to perform well on the \nli in a zero-shot fashion.

This high scores after fine-tuning can be attributed to systems' exploiting the annotation artifacts of data instead of learning to reason with \MCQs.
This claim will be further supported by the \mcqa results.


\begin{table}[t]
    \small
    \centering
    \begin{tabular}{lll}
        \hline \textbf{Model} &\textbf{0-Shot} &\textbf{Tuned}\\ \hline
        RoBERTa-base    & 0.24 & 0.42 \\
        RoBERTa-large   & 0.22 & 0.22 \\
        UnifiedQA-small & \highlight{0.32} & 0.50 \\
        UnifiedQA-base  & 0.23 & 0.59 \\
        UnifiedQA-large & 0.28 & \highlight{0.68} \\
        \hline
        Expert Human & 0.92 & - \\
        Random Baseline & 0.25 & - \\
        \hline
    \end{tabular}
    \caption{Accuracy results of SOTA systems on \mcqa task based on \CQ. Best values are \highlight{highlighted}.}
    \label{tab:MCQA-results}
    \vspace{-1em}
\end{table}

\emph{(2) \mcqa Results}
The \mcqa has all the intricacies of the original \MCQ data absent from the simple annotation artifacts that make it a better alternative to evaluate systems.
As presented in Table~\ref{tab:MCQA-results}, there is a significant gap between the ideal and machine performance in the \mcqa benchmark that further supports the novelty of \CQ and tasks stemming from it.

After investigating the answers, we observe that even the promising large models tend to confuse the enabling v.s. disabling cases.
For example the \emph{UnifiedQA-Large} model, mistakenly chooses a disabling response ``Your car is out of fuel'' for the enabling question ``Gas are typically used for providing energy.\ What makes this possible?''.
This might be explained by the \fact that LMs tend to focus more on correlation of lexical occurrences and statistical patterns (e.g., gas and car/fuel), rather than the actual question.
In addition, similar to \citet{axioms}, we observe that LMs lack understanding of linguistic permutations like negations, and lean toward positive words.

\emph{(3) \gen Results}
As summarized in Table~\ref{tab:gen-results-bl}, the automatic evaluation results, BLEU and ROUGE, are close to zero for all models.
This shows that the models fall short in generating similar to reference precondition even after fine-tuning. 
On the other hand, the human annotation sheds more light on the results and show the relative comparison of the models.

Here the automatic evaluation methods do not sufficiently distinguish between the models as the difference among them are negligible. 
Hence, the comparison rather provides complementary insights to the two discriminative tasks.
This is consistent with similar generation tasks~\cite{rudinger2020thinking}, due to the small number of reference responses and relatively large space of correct responses that makes automatic evaluation of such machine responses an unresolved problem~\cite{Chen2020MOCHAAD}.


\begin{table}[t]
    \small
    \centering
    \begin{tabular}{l|cc|c|c}
        \hline 
        \multirow{2}{*}{\textbf{Model}} 
        &\multicolumn{2}{c|}{\textbf{BLEU}} 
        &\textbf{ROUGE}  
        &\textbf{HUM}\\
        \textbf{} 
        &\textbf{0-Shot} &\textbf{Tuned} 
        &\textbf{Tuned} 
        &\textbf{Info.}\\ \hline
        UnifiedQA-small  & 0.007 & 0.157 & 0.064 & 0.12 \\
        UnifiedQA-base  & 0.006 & 0.303 & 0.115 & 0.28 \\
        UnifiedQA-large  & 0.029 & \highlight{0.330} & 0.128 & \highlight{0.48} \\
        BART-base & 0.046 & 0.091 & \highlight{0.140} & 0.19 \\
        BART-large & 0.041 & 0.058 & 0.117 & 0.11 \\
        GPT2    & \highlight{0.097} & 0.133 & 0.067 & 0.36 \\
        \hline
        Expert Human & - & - & - & 1.0 \\
        \hline
    \end{tabular}
    
    \caption{BLEU-2, ROUGE-2, and human evaluation Information score for results of SOTA systems on the \gen task. Zero-shot ROUGE scores are omitted to save space as they are negligible and do not add additional insight beyond the zero-shot BLEU-2. Best values are \highlight{highlighted}.}
    \label{tab:gen-results-bl}
    \vspace{-1em}
\end{table}

Upon analyzing the results we noticed several patterns in the generated responses. 
First, models tend to generate simple answers mostly discussing the existence or availability of the subject. 
For example, \emph{BART-base} frequently generated patterns such as \QT{ \textless head\textgreater~ is closed} or \QT{You have \textless head\textgreater} some of which were informative.
Second, similar to the \mcqa task, the models tend to confuse enabling and disabling \MCQs. 
For example, \emph{BART-large} generated the enabling \MCQ \QT{The clothes are dirty} instead of disabling \MCQ for the \fact \QT{Washing clothes are used for making fresh again}.

\subsection{Diving in the Tuning Process}
\label{sec:curve-results}
In the above evaluation on \nli, we observe that all models get higher scores after fine-tuning.
Here, we investigate the fine-tuning process to find at what point the model understands the requirements of the task.

\paragraph{Experimental Setup}
We focus on the \emph{RoBERTa-large-MNLI}~\cite{liu2019roberta} model in the \nli task. 
The experimental setup is similar to section~\ref{subsubsec:base-results-setup}.
We evaluate the model's performance on the test split of \nli in checkpoints during the tuning process instead of just at the end of it.
Checkpoints are based on the amount of tuning data the model has observed~($10\%, 20\%, \cdots, 100\%$).

\paragraph{Results}
\begin{figure}[h]
    \centering
    \includegraphics[width=0.7\linewidth]{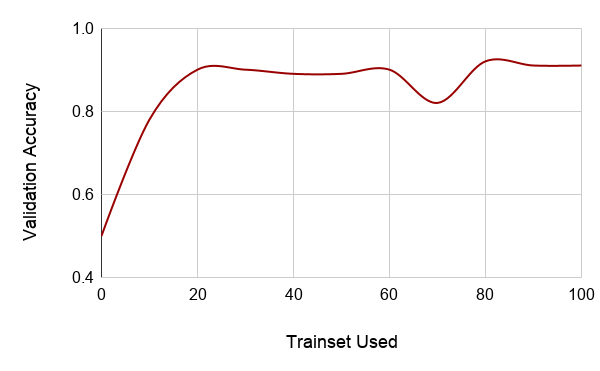}
    \vspace{-1em}
    \caption{F1-Score of fine-tuning \emph{RoBERTa-large-MNLI} with increasing amounts of training~(tuning) data from \nli.
    }
    \label{fig:nli-l-curve}
    \vspace{-1em}
\end{figure}

Figure~\ref{fig:nli-l-curve} plots the changes of score of the model as it gets more tuning data.
The slow saturation of the F1 score here suggests that the instances in \nli are not trivial for the model and it actually has to see a lot of instances to be able to perform the task.
Considering that the \emph{RoBERTa-large-MNLI} has been pre-trained on a vast corpus, our result shows the novelty and uniqueness of the \CQ data.

\subsection{Discussion on Different Relation Types}
\label{sec:other-ralation-results}
Given that \CQ consists of three relations types, we next pose the question of how well the LMs can handle each relation type. 
Here, we break down the results presented in Section~\ref{subsec:base-results} per relation type and discuss the model performance on each type.

\paragraph{Experimental Setup}
Due to simplicity of automatic evaluation, we on focus on the two discriminative tasks, \nli and \mcqa. 
The experimental setup here is similar to section~\ref{subsubsec:base-results-setup}, except that for both zero-shot and fine-tuned settings where we measure the dissected results based on the relation types as well as their aggregation.

\paragraph{Results}
On the \nli task, similar to the challenges for human annotators (Section~\ref{subsec:datacollection}), all NLI models tend to get lower accuracy on instances derived from \emph{Causes} and \emph{Desires} relations, compared to \emph{Usedfor}.
For instance, the \emph{DeBERTa-large-MNLI}, has a 6\% gap between the performance on \emph{UsedFor} and \emph{Causes} instances.
In the \mcqa task, we observe a similar pattern between \emph{Causes} and \emph{Desires} relations on one hand, and \emph{Usedfor} on the other hand.
For instance, the \emph{UnifiedQA-large} mode shows a 13\% gap between instances with \emph{Usedfor} and \emph{Desires}
relations.
The detailed \nli and \mcqa performance results dissected based on relation types are provided in Tables~\ref{tab:NLI-relation-results} and ~\ref{tab:MCQA-relation-results} in the Appendix section.

\subsection{Hard and Soft Preconditions}
\label{sec:soft-hard-results}
In this work, we argued for the use of hard \MCQs as opposed to soft \MCQs used in previous works.
Although semantically different, one may argue that using soft \MCQs may help the models learn the task of reasoning with \MCQs with already existing data.
In this section we test this hypothesis.

\paragraph{Experimental Setup}
Using the approach presented in Section~\ref{sec:benchmark_design}, we created an NLI resource from two available resources with soft \MCQs: ~\citet{rudinger2020thinking} and ATOMIC2020~\cite{hwang2020comet} (Details in Appendix~\ref{subsec:appendix-transfer}).
We focused on the \emph{RoBERTa-large-MNLI}~\cite{liu2019roberta} model, fine-tuned in on the two resources, and evaluate on the test set of \nli.
The experimental setup here is similar to Section~\ref{subsubsec:base-results-setup}.



\paragraph{Results}
Although these resources have an order of magnitude more data (88K instances in ATOMIC2020~\cite{hwang2020comet} and 236K instances in \citet{rudinger2020thinking}), there is more than 10\% gap between the performance of the model tuned on them in the \nli task compared to a model exposed to \CQ data. 
Table~\ref{tab:NLI-transfer}, presents the detailed results of tuning \emph{RoBERTa-large-MNLI} model on each of the NLI-style datasets, while being evaluated on \nli's test subset. 

\section{Related Work}
\label{sec:related_work}

\paragraph{Resources of Preconditions.}
A few resources have provided representations for preconditions of \facts.
\CN~\cite{speer2016conceptnet}'s \textit{HasPrerequisite} relation,
ATOMIC~\cite{sap2019atomic}'s \textit{xNeed} relation, and CauseNet~\cite{heindorf2020causenet} data can express concept dependencies, such as, e.g., before one bakes bread, they need to buy ingredients and go to a store.
Instead of adding new edges, our work annotates existing edges with contextual preconditions, which helps reasoners understand when to use an edge and when not to.
ASER~\cite{zhang2020aser} and ASCENT~\cite{nguyen2020advanced} extract edges from unstructured text together with their associated context.
As such, their knowledge is restricted by information available in text, and they do not express \emph{disabling} \MCQs.
It is also unclear to which extent their contextual edges express \emph{enabling} preconditions, rather than coincidental information.
GLUCOSE~\cite{mostafazadeh2020glucose} comes closer to our work, as they also extract \emph{enabling} \MCQs (e.g., \emph{Possession state that enables X}) via crowdsourcing.
Similarly, PeKo~\cite{kwon-etal-2020-modeling} extract \emph{enabling} \MCQs between event pairs from available text and use it to propose precondition identification and generation tasks between pair of sentences.
However focusing only on causal relations in available text hinders the extent of their tasks.
Both GLUCOSE and PeKo do not explore disabling \MCQs.

\paragraph{Reasoning with Preconditions.}
Few efforts have been made on evaluating \CSadj reasoning with preconditions.
\citet{rudinger2020thinking} focus on modeling weakeners and strengtheners of \CSadj \facts.
Their work adds a \emph{utility} sentence to the \emph{hypothesis-premise} pair in NLI-style tasks and ask whether it weakens or strengthens the relationship of the pair.
Similarly, \citet{hwang2020comet}'s \emph{Hindered by} and \emph{Causes} also focuses on similar relationship for events with focus on presenting a knowledge resource.

Our work differs as we focus on a crisp condition of \emph{enabling/disabling} that can be particularly useful in logic-like reasoning tasks (as opposed to probabilistic inference).
In addition, our task allows the reasoning to be processed as canonical NLI and can benefit from existing NLI architectures instead of modifying them.

\section{Conclusions and Future Work}
\label{sec:conclusion}
We presented, \CQ, a dataset of $12.4$K collected enabling and disabling preconditions of everyday commonsense statements from \CN.
We utilize this resource to create three tasks for evaluating the ability of systems to reason over circumstantial preconditions, namely: \nli, \mcqa, and \gen.
Our evaluation shows that SOTA reasoners largely fall behind human performance, indicating the need for further investigation to develop precondition-aware systems.

Future work should cover the inclusion of preconditions in logical reasoning of the neuro-symbolic reasoners.
It should also expand to multimodal setup or investigate using weak-supervision to gather preconditions.
Alternatively, we can leverage the contributed resource to develop generative models for automated context-aware knowledge base construction~\cite{sorokin2017context}.
\nocite{tandon2019wiqa}

\section*{Ethical Statement}
\label{sec:ethical-considerations}
Though we may present this as we started from openly available data that is both crowdsource-contributed and neutralized, however it still may reflect human biases~\cite{mehrabi2021lawyers}.

During our data collection we did not collect any sensitive information, such as demographic or identity characteristics.
We only limited the annotators to English-speaking users from mainly English-speaking countries such as US, which may add cultural bias to the data.
However, neither our crowd annotators or the expert annotators noticed any offensive language in the questions or the responses.

Given the urgency of addressing climate change we have reported the detailed model sizes and runtime associated with all the experiments in Appendix~\ref{sec:model-sizes-and-run-times}.

\section*{Limitations}
The current \paco still has limitations in the breadth and diversity of preconditions associated with commonsense knowledge. 
However, with more resources we would easily extend the benchmark in both directions to have PaCo v2.0. 
From the breadth perspective, \paco utilizes \CN as source of \CSn statements which has a bounded scope of coverage on commonsense scenarios, even though, to the best of our knowledge, \CN is so far the largest crowd-verified resource on \CSn knowledge.
From the diversity perspective, \paco currently provides 6 preconditions per statements.
This also limits the comprehensiveness of automatic evaluation for the \gen task, similar to \citet{rudinger2020thinking}, in which a correct answer by the test models may not be in the reference set for it to receive high score. This open problem is addressed specifically in some works, e.g. \citet{Chen2020MOCHAAD}.

\section*{Acknowledgement}
We would like to thank Daniel Schwabe for his insightful comments in our paper. We also want to thank our anonymous reviewers whose comments/suggestions helped improve and clarify this
paper.
This work is supported in part by the DARPA MCS program under Contract No.N660011924033 with the United States Office Of Naval Research, the National Science Foundation of United States Grant IIS 2105329, and a Cisco Research Award.




	\bibliographystyle{acl_natbib}
	\bibliography{references}
	\clearpage
	\setcounter{page}{1}

\appendix


\section{Data Collection Details}
\label{sec:appendix:mturk}
We used Amazon Mechanical Turk (AMT)~\cite{crowston2012amazon} to collect the \CQ.
This enabled us to coordinate the study and access a large pool of English-speaking participants as our study population.
The AMT is especially suitable for this study as it can facilitate accessing a diverse population of participants which is necessary for any notion of \CSadj.
Our study on AMT consists of two parts: a tutorial that also serves as a qualification test and the main survey.
In addition, we implemented two levels of quality control: in the first one we use a response checker code and in the second we use human annotators to ensure only high-quality responses wind up into the final data.

\begin{figure}
    \centering
    \includegraphics[width=\linewidth]{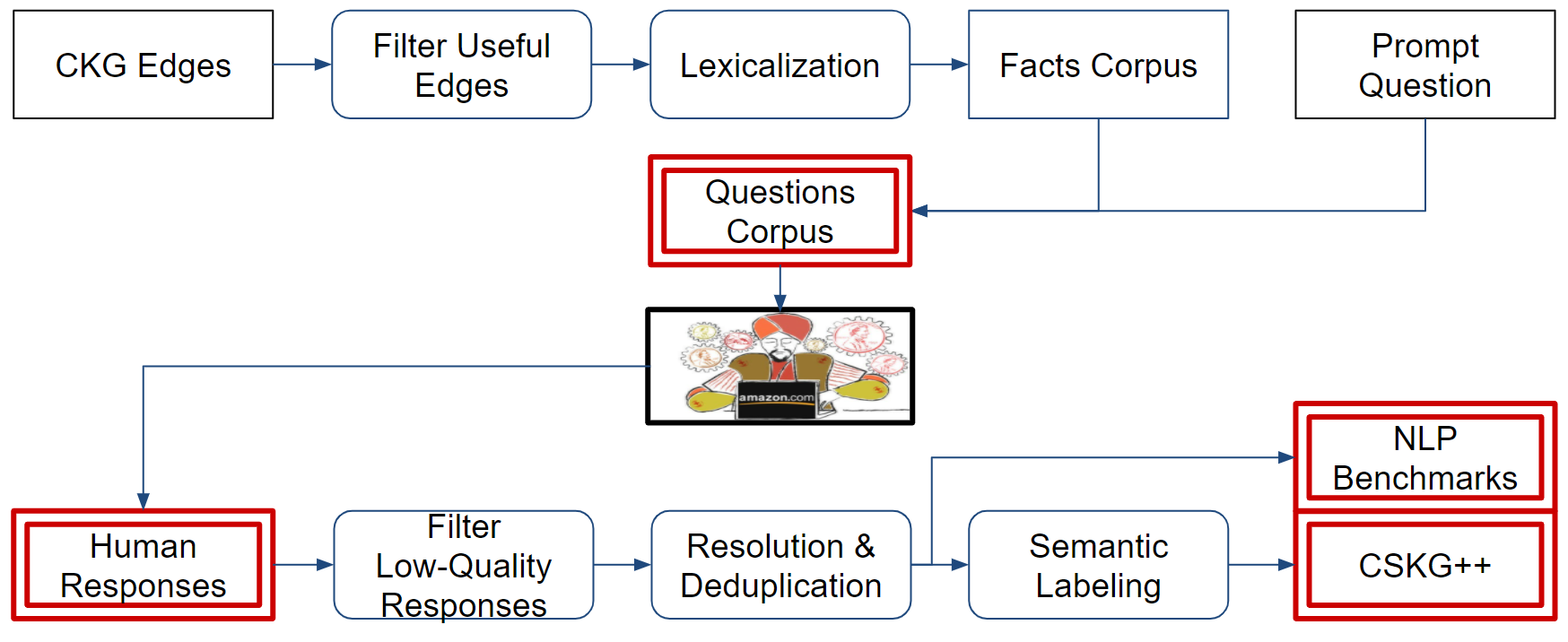}
    \caption{Data-collection and processing in a nutshell}
    \label{fig:data_collection}
\end{figure}

\subsection{Main AMT Survey}
\label{subsec:main-amt-survey}
\begin{figure}
    \centering
    \includegraphics[width=0.8\linewidth]{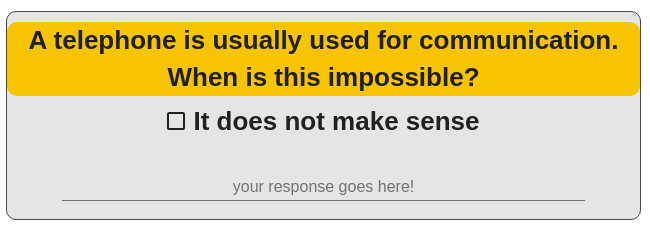}
    \caption{A sample question-unit used in main survey on the AMT}
    \label{fig:question-unit}
\end{figure}

In the main survey, the participants are given a set of question-units (sample in Fig.~\ref{fig:question-unit}) each consists of a factual sentence (discussed in Section~\ref{sec:factual sentences}) followed by a prompt question, then we ask participants to write their responses for each prompt question in the designated text box in front of the unit.
The prompt questions are short questions that ask about the preconditions that enable or disable the factual sentence (e.g. \emph{what makes this possible?}, \emph{when is this impossible}).
The goal of this phase is to use the powers of crowdsourcing to capture as much information as needed to create a dataset of enabling and disabling conditions.


\subsection{Gathering Factual Sentences}
\label{sec:factual sentences}

The first row in Fig.~\ref{fig:data_collection} summarizes the steps to create the factual sentences. Each factual sentence is a short sentence derived from an edge from a \CSadj \ knowledge graph. The information on this knowledge graph is related to everyday situations such as usage of objects (\emph{A net is used for catching fish.}), or capabilities of objects (\emph{Humans are capable of catching a bus.}), etc.~\cite{speer2016conceptnet,ilievski2020consolidating,sap2019atomic}. In our case, the knowledge associated with each factual sentence is extracted from \CN~\cite{speer2016conceptnet}, a well known \CSadj resource. To limit the scope of this work we only focus on \emph{UsedFor}, \emph{Causes}, and \emph{Desires} relations from \CN, however, the method can be extended to any other relation from any other knowledge graph.

To convert the knowledge graph edges to human-readable factual sentences, we used automatic lexicalization methods, similar to~\cite{ma2019towards-hykas,bouraoui2020inducing}. In this method, we define a set of templates to convert the edge to a set of sentence candidates, then use the perplexity score of a language model to pick the best candidate for each edge. The lexicalization is explained in more details in Appendix~\ref{sec:appendix:lexic}.

Since \CN's knowledge is not perfect, some of the generated factual sentences may not fully make sense. Additionally, the automatic conversion of edges to the sentence is not perfect, hence some sentences may have odd grammar (e.g. \emph{An net is used for catch fish}). Consequently, some of the question-units may be hard to understand or just be wrong. To help us find those question-units and ignore them in future iterations, each unit is presented with an  adjacent checkbox labeled \emph{This does not make sense}. The participant may choose to select the checkbox and skip answering that prompt. To make the payment structure fair for the participants, they will get paid regardless of their response.

\subsection{Qualifying Participants}
\label{sec:appendix:quality}
To ensure the participants can understand the task, we prepared detailed instructions that explain to the participants what they need to do and what are the criteria for a good vs bad response. For example, in the instructions, we ask participants to avoid using negative sentences or avoid using pronouns to refer to objects. The instruction is 366 words with an expected reading time of $<5$ mins. Additionally, we have prepared a set of good/bad examples associated with each rule that can also be accessed in the tutorial. Each one of the good/bad examples comes with a short explanation clarifying the reason for its good/bad rating. 

The participants are then asked to take the qualification test as a check on whether they have read and understood the instructions. The qualification test contains 10 multi-choice questions (each with two choices); each containing a question-unit (similar to those that are used in the main survey) with two choices of the possible responses that one may give to them. We have carefully designed each multiple-choice question such that it tests the participants' understanding of the rules individually and give them feedback on their wrong answers. For example, for the rule discouraging the use of negative sentences, we have two questions where the wrong answers contain a negative verb. After successfully passing the test, participants with acceptable scores are granted a qualification badge that allows them to engage in the main survey. It must be noted that the detailed instructions and the good/bad examples are both available in the main survey as a memory refresher for the participants.

For the main survey, we have structured the payment on a per HIT basis, such that the overall compensation be equal to \$15 per hour of work. 
To simplify the annotation process, we grouped 4 statements together in one HIT that helped us reduce the waste time of annotators. 
The participants will be paid by the number of submitted HITs and there will be no min number of HITs for them. 
However, AMT allows us to ban participants that produce low-quality responses from further engaging in our study.
The banned participants were fully compensated for their accepted work (according to automatic evaluation script) up until they are banned.

\subsection{Edge Lexicalization}
\label{sec:appendix:lexic}
Each of the selected edges is lexicalized using a combination of templates and masked LMs described by~\citet{ma2019towards-hykas} and~\citet{bouraoui2020inducing}.
Similar to~\citet{ma2019towards-hykas}, we use a combination of the templates for each relation (e.g. \emph{[subject] is used for [object]}, \emph{[subject] is used by [object]}) and use the perplexity score from the LM to select the best lexicalization for each edge.
However, this method does not guarantee the selection of the best lexicalization as the perplexity score reflects the probability of the sentence tokens appearing in that specific order rather than the sentence's grammatical correctness.
To mitigate this issue, in addition to the above method, following~\cite{bouraoui2020inducing}, we let the LM adjust the templates as well by adding one masked token to some templates (e.g. \emph{[subject] is used [MASK] [object]}) and let the LM fill the \emph{mask} before filling the \emph{subject} and the \emph{object} slots of the template.

\section{Results in More Details}
\label{sec:appendix:results}





\subsection{Edge Selection Results}
\label{sec:appendix:edge}
In this section, we provide further evidence to support the decision to use the \emph{UsedFor} edges without any additional filtering.
First, we showcase the lack of correlation between a hand-annotated usefulness indication of the \MCQ statements and existing quantitative methods/scores. Then, in a similar setup, we show that the \emph{UsedFor} edges have a higher usefulness score.

For the first study, we only focus on \emph{UsedFor} edges. For each metric, we randomly sample 20 edges in each percentile of the metric and hand-annotate the usefulness of sampled edges in each percentile. Then, for each percentile-metric, we report the percentage of edges that were considered useful for our study. The results in Table~\ref{tab:edge_metric_results}, summarizes the usefulness score for three of the percentile buckets for three of the metrics. For the \emph{perplexity} score we used the RoBERTa~\cite{liu2019roberta} language model on the lexicalized edges, for the \emph{Salient} score we used DICE metrics~\cite{chalier2020joint}, and for the \emph{weight} score we use the weights from the \CN~\cite{speer2016conceptnet} itself.
The usefulness scores suggest that a higher score may or may not result in more useful edges which makes using them for filtering edges tricky. This study is by no means conclusive due to both the small sample sizes and a small number of trials, however, it led us to choose the edges solely based on relation type and leave further filterings to future work.

\begin{table}
    \small
    \centering
    \begin{tabular}{llll}
        \hline \textbf{Metric} &\textbf{[0,10](\%)} &\textbf{[50,60](\%)} &\textbf{[90,100](\%)}\\ \hline
            Perp.    &75& 95 & 90  \\
            Salient  &80& 100& 95 \\
            Weight   &95& 90&  90  \\
        \hline
    \end{tabular}
    \caption{hand-annotated usefulness indication of the precondition statements for top/bottom/mid percentile buckets of the quantitative methods. The $[A, B]$ label indicates edges with the metric score in the range of $[A, B]$ percentile of the metric score.}
    \label{tab:edge_metric_results}
\end{table}

For the second study, Table~\ref{tab:edge_relation_results}, we group edges based on their relations only and compute the usefulness score for each relation. The results showed that \emph{UsedFor} edges tend to generally be more useful for our annotation task.
This couple with the \fact that \emph{UsedFor} edges could be annotated with both enabling and disabling preconditions led us to focus on them for this study.

\begin{table}
    \small
    \centering
    \begin{tabular}{ll}
        \hline \textbf{Metric} &\textbf{Score(\%)}\\ \hline
            UsedFor    &95 \\
            CapableOf  &90 \\
            RelatedTo  &40 \\
        \hline
    \end{tabular}
    \caption{hand-annotated usefulness indication of the precondition statements three of the \CN relations}
    \label{tab:edge_relation_results}
\end{table}

\subsection{Additional Results from \nli}
\label{subsec:appendix-NLI}
Table~\ref{tab:appendix:NLI} presents some error cases that each model predicts on the test subset of P-NLI.
\begin{table*}
    \small
    \centering
    \begin{tabular}{p{0.05\linewidth}p{0.48\linewidth}p{0.35\linewidth}p{0.05\linewidth}}
        \hline
        \textbf{Model}
        & \textbf{\textit{\underline{\Fact}}}
        & \textbf{\textit{\underline{Context}}}
        & \textbf{*}
        \\
        \hline
        TE & You can typically use self adhesive label for labelling things & The self adhesive label runs
        out of glue.
        & FP
        \\
        & Acoustic ceiling is typically used for dampening sound.
        & in rooms with noise above a certain decibel.
        & FP
        \\
        & You can typically use self adhesive label for labelling things.
        & Labeling things that are wet.
        & FP
        \\
        & Farm is typically used for raising crops.
        & Enough rain should be available.
        & FN  \\
        roberta & You can typically use pets to provide companionship
        & the pet is dog.
        & FN
        \\
        & Acoustic ceiling is typically used for dampening sound         & The sound is too loud
        & FP
        \\
        \hline
    \end{tabular}
    \caption{Test results of SOTA systems on NLI task based on the \CQ. FP: False Positive, FN: False Negative}
    \label{tab:appendix:NLI}
\end{table*}

As our version of NLI only consists of \emph{Entailment} and \emph{Contradiction} labels, we discuss the results using binary classification terminology.

In addition, the detailed results of Table~\ref{tab:NLI-results} dissected by the relation types are provided in Table~\ref{tab:NLI-relation-results}.

\begin{table}
    \small
    \centering
    \begin{tabular}{llll}
        \hline \textbf{Model} & \textbf{Rel.} &\textbf{0-Shot} &\textbf{Tuned}\\ \hline
        RoBERTa-large-MNLI & UsedFor & 0.34 & 0.85 \\
                        & Causes  & 0.48 & 0.90 \\
                        & Desires & 0.48 & 0.90 \\
                        & All     & 0.47 & 0.90 \\ \hline
        BART-large-MNLI & UsedFor & 0.51 & 0.91 \\
                        & Causes  & 0.41 & 0.82 \\
                        & Desires & 0.46 & 0.89 \\
                        & All     & 0.48 & 0.89 \\ \hline
        DeBERTa-base-MNLI & UsedFor & 0.37 & 0.91 \\
                        & Causes  & 0.32 & 0.84 \\
                        & Desires & 0.38 & 0.88 \\
                        & All     & 0.37 & 0.89 \\ \hline
        DeBERTa-large-MNLI & UsedFor & 0.38 & 0.94 \\
                        & Causes  & 0.31 & 0.88 \\
                        & Desires & 0.36 & 0.90 \\
                        & All     & 0.36 & 0.92 \\ \hline
        DeBERTa-xlarge-MNLI & UsedFor & 0.37 & 0.94 \\
                        & Causes  & 0.31 & 0.88 \\
                        & Desires & 0.37 & 0.89 \\
                        & All     & 0.37 & 0.91 \\
       \hline
    \end{tabular}
    \caption{F1-Macro results of SOTA systems on \nli task based on \CQ dissected based on relation type}
    \label{tab:NLI-relation-results}
\end{table}

\begin{table}
    \small
    \centering
    \begin{tabular}{llll}
        \hline \textbf{Model} & \textbf{Rel.}  &\textbf{0-Shot} &\textbf{Tuned}\\ \hline
        RoBERTa-base    & UsedFor & 0.23 & 0.38 \\
                        & Causes  & 0.21 & 0.41 \\
                        & Desires & 0.27 & 0.38 \\
                        & All     & 0.24 & 0.42 \\ \hline
        RoBERTa-large   & UsedFor & 0.19 & 0.21 \\
                        & Causes  & 0.28 & 0.23 \\
                        & Desires & 0.23 & 0.22 \\
                        & All     & 0.22 & 0.22 \\ \hline
        UnifiedQA-small & UsedFor & 0.37 & 0.55 \\
                        & Causes  & 0.35 & 0.53 \\
                        & Desires & 0.31 & 0.45 \\
                        & All     & 0.32 & 0.50 \\ \hline
        UnifiedQA-base  & UsedFor & 0.56 & 0.67 \\
                        & Causes  & 0.21 & 0.60 \\
                        & Desires & 0.22 & 0.53 \\
                        & All     & 0.23 & 0.59 \\ \hline
        UnifiedQA-large & UsedFor & 0.31 & 0.76 \\
                        & Causes  & 0.26 & 0.68 \\
                        & Desires & 0.26 & 0.61 \\
                        & All     & 0.28 & 0.68 \\
        \hline
    \end{tabular}
    \caption{Accuracy results of SOTA systems on \mcqa task based on \CQ}
    \label{tab:MCQA-relation-results}
\end{table}

\subsection{Details of Soft Preconditions on \nli}
\label{subsec:appendix-transfer}

In order to convert the ATOMIC2020~\cite{hwang2020comet} to an NLI-style task, we method similar to \nli and focused on three relations \emph{HinderedBy}, \emph{Causes}, and \emph{xNeed}. 
From these relations, \emph{HinderedBy} is converted to \emph{Contradiction} and the rest are converted to \emph{Entailment} instances.

For converting \citet{rudinger2020thinking}, we focused on SNLI subset of their data and used the concatenation of SNLI's \QT{Hypothesis} and \QT{Premise} as hypothesis and their \QT{Update} sentence as premise.

Table~\ref{tab:NLI-transfer}, presents the detailed results of tuning \emph{RoBERTa-large-MNLI} model on each of the NLI-style datasets, while being evaluated on \nli's test subset. 
\begin{table}
    \small
    \centering
    \begin{tabular}{lll}
        \hline 
        \textbf{Tune Dataset} & \textbf{Relation} &\textbf{F1-Macro}\\ \hline
        \CQ  
            & UsedFor & 0.85 \\
            & Causes  & 0.90 \\
            & Desires & 0.90 \\ 
            & All     & 0.90 \\
        \hline
        \citet{hwang2020comet}  
            & UsedFor & 0.50 \\
            & Causes  & 0.50 \\
            & Desires & 0.45 \\ 
            & All     & 0.48 \\
        \citet{rudinger2020thinking}  
            & UsedFor & 0.84 \\
            & Causes  & 0.80 \\
            & Desires & 0.82 \\ 
            & All     & 0.83 \\
        \hline
    \end{tabular}
    \caption{Results of RoBERTa-large-MNLI model on test set of \nli after being tuned on different datasets, dissected based on relation type.}
    \label{tab:NLI-transfer}
\end{table}

\section{Model Sizes and Run-times}
\label{sec:model-sizes-and-run-times}

For table \ref{tab:NLI-results}, Runtimes: TE=2hr,rbrta=2.5hr, dbrta-base=0.5hr, dbrta-large=2hr, dbrta-xlarge=3.5hr, BART-large=2hr and \#params: TE=0.5M, rbta=356M, dbrta-base=141M, dbrta-large=401M, dbrta-xlarge=751M, BART-large=407M.  
For table \ref{tab:MCQA-results}, Runtimes:rbta-base=1hr,   rbta-large=2hr,   uqa-small=1hr,   uqa-base=4hr, uqa-large=20hr and \#params: rbta-base=124M,rbta-large=355M,  uqa-small=60M,  uqa-base=222  M,uqa-large=737M.
In table \ref{tab:simple-test}, Runtimes: uqa, gpt2=10min and \#params: gpt2=1.5B.
Finally in table \ref{tab:gen-results-bl}, Runtimes:uqa-small=1hr,    uqa-base=2hr,  uqa-large=6hr, gpt2=1.5B, bart-base=139M, bart-large= and \#params:   uqa-small=60M,uqa-base=222 M, uqa-large=737M, gpt2=1.5B, bart-base=139M, bart-large=406M.

\end{document}